\pdfoutput=1

\documentclass[11pt]{article}

\usepackage[preprint]{acl}

\usepackage{amsmath,amsfonts,bm}

\def\eqref#1{equation~\ref{#1}}

\def\1{\bm{1}}

\DeclareMathAlphabet{\mathsfit}{\encodingdefault}{\sfdefault}{m}{sl}
\SetMathAlphabet{\mathsfit}{bold}{\encodingdefault}{\sfdefault}{bx}{n}

\def\gX{{\mathcal{X}}}
\def\gY{{\mathcal{Y}}}

\newcommand{\E}{\mathbb{E}}

\newcommand{\R}{\mathbb{R}}

\DeclareMathOperator*{\argmin}{arg\,min}

\usepackage[group-separator={,}]{siunitx}
\usepackage{booktabs} 
\usepackage{color}
\usepackage{listings}
\usepackage{enumitem}
\usepackage{threeparttable}
\usepackage{multirow}
\usepackage{hhline}
\usepackage{subcaption}
\usepackage{array}
\usepackage{pifont}
\usepackage{booktabs}
\usepackage{dsfont}
\usepackage{microtype}
\usepackage{graphicx}
\usepackage{stmaryrd}
\usepackage{amsmath}

\usepackage{amssymb}
\usepackage{amsthm}
\usepackage{url}
\usepackage{bbm}
\usepackage{xspace}
\usepackage[normalem]{ulem}
\usepackage{booktabs, siunitx}
\usepackage{xcolor}
\usepackage{colortbl}
\usepackage{tkz-euclide}
\usepackage[framemethod=tikz]{mdframed}
\usepackage{empheq}
\usepackage[most]{tcolorbox}
\usepackage{adjustbox}
\usepackage{wrapfig}
\usepackage[capitalize]{cleveref}
\usepackage{algorithm}
\usepackage{algorithmic}
\usepackage{pgfplots}
\usepgfplotslibrary{groupplots}
\usepackage{pgfplotstable}
\usepackage{bm}
\usepackage{geometry}
\usepackage{CJKutf8}
\usepackage[encapsulated]{CJK}

\usepackage{balance}
\expandafter\def\expandafter\UrlBreaks\expandafter{\UrlBreaks
  \do\a\do\b\do\c\do\d\do\e\do\f\do\g\do\h\do\i\do\j%
  \do\k\do\l\do\m\do\n\do\o\do\p\do\q\do\r\do\s\do\t%
  \do\u\do\v\do\w\do\x\do\y\do\z\do\A\do\B\do\C\do\D%
  \do\E\do\F\do\G\do\H\do\I\do\J\do\K\do\L\do\M\do\N%
  \do\O\do\P\do\Q\do\R\do\S\do\T\do\U\do\V\do\W\do\X%
  \do\Y\do\Z}

\newtheorem{theorem}{Theorem}[section]

\newtheorem{example}{Example}[section]

\newcommand{\ARC}[0]{ARC\xspace}
\newcommand{\ARCT}[0]{ARC-Tran\xspace}

\def\T{{\scriptscriptstyle\mathsf{T}}}

\newcommand{\oracle}[0]{exhaustive}
\newcommand{\Oracle}[0]{Exhaustive}

\newcommand{\nnmodel}[0]{F}
\newcommand{\params}[0]{\theta}

\newcommand{\Tinsword}{T_{\emph{Dup}}}

\newcommand{\Tsubword}{T_{\emph{SubSyn}}}

\DeclareMathOperator{\bfx}{{\mathbf{x}}}

\DeclareMathOperator{\bfz}{{\mathbf{z}}}

\newcommand{\dataset}{D}

\newcommand{\alphabet}{\Sigma}
\newcommand{\spec}{S}

\def\substring#1#2#3{#1_{#2:#3}}

\newcommand{\Exhaustiveshort}{EX Acc.}
\newcommand{\Verifiedshort}{CF Acc.}

\newcommand{\lstm}{\textsc{lstm}}
\newcommand{\trans}{\textsc{trans}}

\newcommand{\ibp}[1]{\widehat{#1}}

\renewcommand{\geq}{\geqslant}

\newcommand{\len}[1]{\textsc{len}_{#1}}

\newcommand{\x}{\bfx}

\usepackage{times}
\usepackage{latexsym}

\usepackage[T1]{fontenc}

\usepackage[utf8]{inputenc}

\usepackage{microtype}

\usepackage{inconsolata}

\usepackage{graphicx}

\title{A One-Layer Decoder-Only Transformer is a Two-Layer RNN:\\With an Application to Certified Robustness}

\author{Yuhao Zhang \and Aws Albarghouthi \and Loris D'Antoni \\
University of Wisconsin-Madison \\
\small{
   \textbf{Correspondence:} \href{mailto:yuhaoz1997@gmail.com}{yuhaoz1997@gmail.com}
 }}
  
\begin{document}
\maketitle
\begin{abstract}
This paper reveals a key insight that \emph{a one-layer decoder-only Transformer is equivalent to a two-layer Recurrent Neural Network (RNN)}.
Building on this insight, we propose \ARCT, a novel approach for verifying the robustness of decoder-only Transformers against arbitrary perturbation spaces.
Compared to \ARCT, current robustness verification techniques are \emph{limited} either to specific and length-preserving perturbations like word substitutions or to recursive models like LSTMs.
\ARCT addresses these limitations by meticulously managing position encoding to prevent mismatches and by utilizing our key insight to achieve precise and scalable verification.
Our evaluation shows that \ARCT (1) trains models more robust to arbitrary perturbation spaces than those produced by existing techniques and (2) shows high certification accuracy of the resulting models.
\end{abstract}

\section{Introduction}
Large language models (LLM) have proven incredibly powerful in a vast range of tasks~\citep{chatgpt, gpt3, bert}.
The effectiveness of LLMs is primarily attributed to the Transformer architecture, which is the backbone of many leading models and is believed superior to earlier architectures such as RNNs and LSTMs.

\begin{figure}[t]
    \centering
    \subfloat[A one-layer decoder-only Transformer]{
    \includegraphics[width=0.8\linewidth]{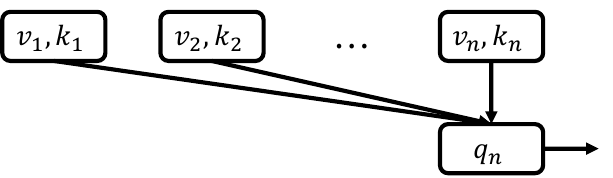}
    }
    \hfill
    \subfloat[A two-layer RNN]{
    \includegraphics[width=0.8\linewidth]{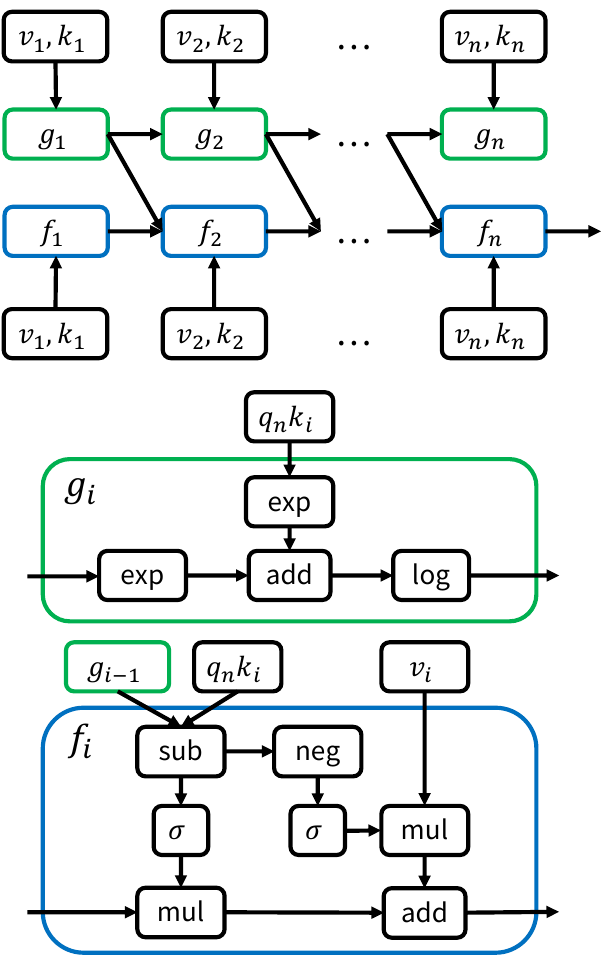}
    }
    \caption{A one-layer decoder-only Transformer
is equivalent to a two-layer RNN.}
    \label{fig: illu_rewrite}
    \vspace{-1.5em}
\end{figure}

There has been a notable shift in Transformer architectures from bidirectional encoder structures like BERT~\citep{bert} to decoder-only structures like GPT-3~\citep{gpt3}.
This shift is driven by the excellent performance of decoder-only models in various generative tasks.
In this paper, we highlight a key insight:
\begin{center}
    \emph{A one-layer decoder-only Transformer
is equivalent to a two-layer RNN.}
\end{center}
We illustrate this insight in \Cref{fig: illu_rewrite} and provide a formal proof in \Cref{sec: observation}. 
We believe this insight opens up promising future research directions, and we present its application to certified robustness.

\paragraph{Application to Certified Robustness}
Despite their impressive performance on normal inputs, large language models are susceptible to adversarial examples that are small input perturbations, causing the model's accuracy and generation quality to plummet~\citep{llmrobustness, zhang2024codefort}.
Techniques such as convex over-approximation~\citep{crown, deeppoly} and randomized smoothing~\cite{randomizesmoothing} have been developed to verify the robustness of neural networks. 
These techniques can prove the robustness of a model by showing the absence of adversarial examples near a given input.

This paper focuses on the robustness of the decoder-only Transformer architecture, which is fundamental to many advanced LLMs.
Existing work~\citep{transformer_lp_1, transformer_lp_2} for verifying transformer robustness primarily addresses $l_p$ norm perturbations of embeddings, which only captures word substitutions.
However, ensuring language model robustness requires addressing a wider range of transformations like insertions, deletions, substitutions, swaps, and their combinations~\citep{arc, a3t, text_crs}. 
ARC~\citep{arc} is the only technique that verifies model robustness against arbitrary perturbation spaces, but it only works on recursive models like LSTMs.
Therefore, bridging this gap of robustness verification of decoder-only Transformers against arbitrary perturbation spaces is a critical step toward the trustworthiness of LLMs.


We identify two challenges in verifying decoder-only Transformers, discussed in \Cref{sec: approach}, and propose \ARCT as a solution. 
The first challenge involves position encoding, which can complicate verification when perturbed inputs vary in length. 
\ARCT addresses this by introducing a new embedding strategy, which adjusts for changes in input length when integrated with \ARC~\citep{arc}. 
The second challenge is the dependency of the hidden state in decoder-only Transformers on \emph{all} preceding states, unlike in LSTMs where the dependency is only on the immediate previous state. 
This dependency complicates the trade-off between precision and scalability in verification. 
\ARCT addresses this by reinterpreting the attention mechanism in a one-layer decoder-only Transformer as a two-layer RNN, as shown in \Cref{fig: illu_rewrite}.

\paragraph{Contributions}
We make the following contributions: 
(1) We present a key insight that a one-layer decoder-only Transformer is equivalent to a two-layer RNN, enabling the efficient and precise abstract interpretation of decoder-only Transformers against programmable perturbation spaces.
(2) We present \ARCT, an approach for verifying and training certifiably robust decoder-only Transformers.
(3) Our evaluation shows that \ARCT trains models more robust against arbitrary perturbation spaces than those produced by existing techniques; \ARCT shows high certification accuracy of the resulting models.


\section{A Key Insight}
\label{sec: observation}
This section presents the key insight that a one-layer decoder-only Transformer is equivalent to a two-layer RNN.

We begin by defining the $n$th transformer state, computed using the key, query, and value vectors $\mathbf{k}, \mathbf{q}, \mathbf{v}$, as follows,
\begin{align}
    \sum_{i=1}^n v_{i} \frac{\mathrm{e}^{q_nk_i}}{\sum_{j=1}^n \mathrm{e}^{q_nk_j}}.\label{eq: def_trans_raw}
\end{align}

Next, denoting the sigmoid function as $\sigma$, we express the relation of $f_i$ and $g_i$ in the context of an RNN:
    \begin{align}
        f_i &= \begin{cases}
            v_i\sigma(q_nk_i - g_{i-1}) \\
            ~~~~+ f_{i-1}\sigma(g_{i-1} - q_nk_i), & i > 1 \\
            v_1, & i=1
        \end{cases} \label{eq: rnn_trans} \\
        g_i &= \begin{cases} 
        \log(\mathrm{e}^{g_{i-1}} + \mathrm{e}^{q_nk_i}), & i > 1 \\
        q_nk_1,& i=1
        \end{cases}. \label{eq: rnn_trans_g} 
    \end{align}

\begin{theorem}\label{theorem: rewrite}
    $f_n$ in \Cref{eq: rnn_trans} is equivalent to \Cref{eq: def_trans_raw}.
\end{theorem}
We provide the proof of \Cref{theorem: rewrite} in the Appendix and the key to the proof is rewriting the Softmax function as a series of Sigmoid functions.

For numerical stability, we further note that \Cref{eq: rnn_trans_g} needs to be implemented as: 
\[g_i = \max(g_{i-1}, q_nk_i) + \mathrm{log1p}(e^{-|g_{i-1} - q_nk_i|}).\]

\section{Problem Definition}

We consider a classification setting with a neural network $F_\theta$ with parameters $\theta$, trained on samples from domain $\gX$ and labels from $\gY$.
The 
 domain $\gX$ is a set of  strings over a finite set of vocabulary $\alphabet$, i.e., $\gX = \alphabet^*$.
We use $\bfx \in \alphabet^*$ to denote a string; $x_i \in \alphabet$ to denote the $i$th element of the string; $\substring{\bfx}{i}{j}$ to denote the substring $x_i, \ldots, x_{j}$; and $\len{\bfx}$ to denote the length of the string.

\paragraph{Programmable perturbation space} A \emph{perturbation space} $S$ is a function takes a string $\bfx$ and returns a set of possible perturbed strings obtained by modifying $\bfx$.
Intuitively, $S(\bfx)$ denotes a set of strings
that are semantically \emph{similar} to $\bfx$ and therefore
should receive the same prediction.
We follow \citet{a3t, arc}
to define $S$ \emph{programmatically}
as a set of string transformations.

A string transformation $T$ is a pair $(\varphi, f)$, where 
$\varphi$ is the \emph{match} function, a Boolean function that specifies the substrings to which the transformation can be applied; and
$f$ is the \emph{replace} function, which specifies
how the substrings matched by $\varphi$ can be replaced.

\begin{example}
\label{exp:transformation}
Let $T_\emph{del}$ be a string transformation that deletes the stop words ``\emph{to}'' and ``\emph{the}''. 
Formally, $T_\emph{del}=(\varphi_\emph{del},f_\emph{del})$, where $\varphi_\emph{del}$ and $f_\emph{del}$ are:
\vskip -0.1in
\begin{small}
\begin{align*}
    \varphi_\emph{del}(x) = \begin{cases}
    1,& x\in \{\text{``to'', ``the''}\} \\
    0,& \text{otherwise}
\end{cases}, \quad f_\emph{del}(x) = \{\epsilon\},
\end{align*}
\end{small}%

Let $T_\emph{sub}$ be a  transformation substituting the word ``\emph{movie}'' with ``\emph{movies}'' or ``\emph{film}''. Formally,
$T_\emph{sub}=(\varphi_\emph{sub},f_\emph{sub})$, where $\varphi_\emph{sub}$ and $f_\emph{sub}$ are:
\vskip -0.1in
\begin{small}
\begin{align*}
    \varphi_\emph{sub}(x) = \begin{cases}
    1,& x = \emph{``movie''}\\
    0,& \text{otherwise}
\end{cases}, f_\emph{sub}(x) = \left\{\!\begin{aligned}
&~~\emph{``film''},\\
&\emph{``movies''}
\end{aligned}\right\}
\end{align*}
\end{small}%
\end{example}

We can compose different string transformation to construct perturbation space $S$:
\begin{align}
    \spec=\{(T_1, \delta_1), \dots ,(T_n, \delta_n)\}, \label{eq:progspec}
\end{align}
where each $T_i$ denotes a string transformation that can be applied \emph{up to} $\delta_i \in \mathbb{N}$ times. 
The detailed definition of programmable perturbation space can be found in \citet{a3t, arc}.
We then illustrate with an example.

\begin{example}
\label{exp: perturbspace}
Let $\spec =\{(T_\emph{del},1),(T_\emph{sub},1)\}$ be a perturbation space that applies $T_\emph{del}$ and $T_\emph{sub}$ to the given input sequence up to once each.
If $\bfx=$``\emph{to the movie}'', the perturbation space $\spec(\bfx)$ is $\{\text{``\emph{to the movie}''}, \text{``\emph{the movie}''}, \text{``\emph{to the film}''}, \\\text{``\emph{to movies}''}, \ldots\}$
\end{example}

\paragraph{Robustness to programmable perturbation space}
Given a string $\bfx$ with label $y$
and a programmable perturbation space $S$,
We say that a neural network $F_\theta$ is \textit{$S$-robust} on $(\bfx,y)$ iff
\begin{align}
    \forall \bfz \in S(\bfx) \ldotp F_\theta(\bfz) = y \label{eq:robustness}
\end{align}

We aim to \emph{certify}, or prove, $S$-robustness (\Cref{eq:robustness}) of the neural network for a pair $(\bfx,y)$.
Given a certification approach, we can then use it within an adversarial training loop to yield certifiably robust networks.

\paragraph{Robustness certification}
We will  certify $S$-robustness by solving
an \emph{adversarial loss} objective: 
\begin{align} 
\max_{\bfz \in S(\bfx)} L_\theta(\bfz,y)\label{eq:advloss}
\end{align} 
where we assume that the loss function $L_\theta$ is $< 0$ when $F_\theta(\bfz) = y$
and $\geq 0$ when $F_\theta(\bfz) \neq y$.
Therefore, if we can show that the solution to the above problem is $< 0$, then we have a certificate of $S$-robustness.

\paragraph{Certified training}
If we have a procedure to compute adversarial loss,
we can use it for \emph{adversarial training} by solving the following
\emph{robust optimization} objective~\citep{pgd},
where $\mathcal{D}$ is the data distribution:
\begin{align} 
\argmin_\theta \mathop{\mathbb{E}}_{(\bfx,y) \sim \mathcal{D}} \left[\max_{\bfz \in S(\bfx)} L_\theta(\bfz,y)\right]\label{eq:advlosstrain}
\end{align} 
\section{Challenges and Approaches}
\label{sec: approach}
In this section, we point out two challenges in verifying decoder-only Transformers and propose \ARCT with corresponding approaches to overcome these challenges.

\subsection{Position Encoding under Arbitrary Perturbation Spaces}

Position encoding is a crucial component in the Transformer architecture. It allows the model to incorporate positional information of tokens in a sequence. Since the Transformer relies solely on attention mechanisms and has no inherent token order notion, position encodings are added to the input embeddings to inject positional information. 

Position encodings do not cause an issue for previous work~\citep{transformer_lp_1, transformer_lp_2} because they only consider length-preserving perturbation spaces, such as synonym substitutions. However, regarding arbitrary programmable perturbation spaces, perturbed inputs can have different lengths. This difference in lengths necessitates careful handling of the embeddings before attention computation.
Conventionally, before being fed to the attention computation, the position encoding $p_i$ has already been added to the embedding of the $i$th token, i.e., $e_i = t_i + p_i$, where $t_i$ is the token embedding. Directly applying \ARC~\citep{arc} to $e_i$ can cause position mismatches.
To solve this issue, we introduce a new embedding $e_{i,j}$, which represents the original $j$th input token being moved to the $i$th position in the perturbed input, denoted as $e_{i,j}=t_j+p_i$. 
Intuitively, the new embedding separates the token embedding and position encoding and adapts to the input-length change.

Let $G_{i,j}^S$ capture the set of hidden states that have all programmable perturbation space $S$ applied and have the $i$th token in the perturbed input and the $j$th token in the original input (see Eq~6~and~13 in \citet{arc}), the following example illustrates how to overcome the position encoding challenge using the new embedding $e_{i,j}$, assuming we still use an LSTM in addition to position encodings instead of an decoder-only Transformer, which will be explained in the next challenge.
\begin{example}
\label{ex: embedding}
Let $\bfx = \emph{``to the''}$, $t_1$ and $t_2$ be the token embeddings for the first and second word, respectively, and $p_1$ and $p_2$ be the corresponding position encodings.
Conventionally, the embeddings fed to the next LSTM layer (or Transformer layer) are computed as $e_1 = t_1+p_1$ and $e_2=t_2+p_2$.
As a result, $G_{1,2}^{\{(T_\emph{del}, 1)\}}$ will be computed as $\{\lstm(e_1,h_0), \lstm(e_2,h_0)\}$.
However, the second state in the above computation is incorrect because after deleting ``to'', the word ``the'' becomes the first word in the sentence, and its embedding should be $t_2+p_1$ instead of $e_2=t_2+p_2$.

In contrast, leveraging the new embedding $e_{i,j}$, we can compute the correct state set $G_{1,2}^{\{(T_\emph{del}, 1)\}} = \{\lstm(e_{1,1},h_0), \lstm(e_{1,2},h_0)\}$.

\end{example}

\subsection{Handling Decoder-Only Transformer} 
The key difference between an LSTM and a decoder-only transformer is that for $i$th state, an LSTM takes the previous state as its input.
In contrast, an decoder-only transformer takes all the previous states as inputs because it needs to compute the attention over all previous states.

A straightforward approach is storing all previous states instead of a single previous state during computation. 
We first discuss which states need to be stored by investigating the computation of the attention mechanism in Transformers.
Formally, we define the $n$th transformer state on the perturbed input $\bfz$ as
\begin{align}
    \trans(\bfz_{1:n},m) = \sum_{i=1}^n v_{i} \frac{\mathrm{e}^{q_nk_i}}{\sum_{j=1}^n \mathrm{e}^{q_nk_j}},\label{eq: def_trans}
\end{align}
where $m$ is the index mapping from $\bfz$ to the original input $\bfx$, and $q_n=W_qe_{n,m(n)}$, $k_i=W_ke_{i,m(i)}$, $v_i=W_ve_{i,m(i)}$ are query, key, value vectors computed from the new embedding $e_{i,m(i)}$.
From \Cref{eq: def_trans}, we need to take care of $\bfz$, $m$, $q_{1:i}$, $k_{1:i}$, and $v_{1:i}$.
Among these, the index mapping $m$ and the perturbed input $\bfz$ will be maintained by the definition of $G_{i,j}^S$ in \ARC, as shown in Example~\ref{ex: embedding}.
Therefore, we need to store the remaining three matrices (or three lists of vectors) $q_{1:i}$, $k_{1:i}$, $v_{1:i}$ for computing state $ \trans(\bfz_{1:n},m)$.
However, this approach has two drawbacks.
First, it needs an extra $O(\len{\bfx})$ factor on memory consumption for Transformers compared to \ARC for LSTMs.
Second, the interval abstraction used in \ARC will cause additional over-approximation when abstracting $q_{1:i}$, $k_{1:i}$, and $v_{1:i}$ (see the next example). 

\begin{example}
\label{ex: trans_approximate}
    Consider $\bfx = \emph{``A B''}$, where each word has one synonym (``a'', ``b'', respectively) that can be substituted. Let $e_{i,i}$ be the embedding of the $i$-th original token, and $e^s_{i,i}$ be the embedding of the $i$-th substituted token.
Without loss of generality, we only consider the abstraction of $k_{1:i}$ in this example.
    Let $\ibp{A}_{i,j}^{S}$ denote the abstraction of $k_{1:i}$ where the perturbed prefixes have had all transformation in a space $S$ applied on the original prefix $\bfx_{1:j}$.
    We have $\ibp{A}_{1,1}^{\varnothing} = \alpha(\{W_ke_{1,1}\})$, $\ibp{A}_{1,1}^{\{(\Tsubword{},1)\}} = \alpha(\{W_ke^s_{1,1}\})$.
    At the first state, the interval abstractions are still tight.

    At the second state, let $[a;b]$ denote the concatenation of vectors $a$ and $b$. Then, we have 
    \begin{align}
        \ibp{A}_{2,2}^{\{(\Tsubword{},1)\}} &= \alpha\left(\begin{tabular}{c}
          $[\ibp{A}_{1,1}^{\varnothing} ; W_ke^s_{2,2}],$ \\
          $[\ibp{A}_{1,1}^{\{(\Tsubword{},1)\}}; W_ke_{2,2}]$
        \end{tabular}\right) \nonumber \\
        &= [\alpha(\{W_ke_{1,1}, W_ke^s_{1,1}\}) ; \nonumber\\ 
        &~~~~~~\alpha(\{W_ke_{2,2}, W_ke^s_{2,2}\})]
    \end{align}
    The $\ibp{A}_{2,2}^{\{(\Tsubword{},1)\}}$ computed above introduces an extra over-approximation, which \ARC will not introduce for LSTMs
    , as $\ibp{A}_{2,2}^{\{(\Tsubword{},1)\}}$ contains a perturbed input ``a b'', which substituted two words with their synonyms.
\end{example}

    The over-approximation introduced in Example~\ref{ex: trans_approximate} is due to the fact that \ARC is only able to record the previous string transformation, which is sufficient for LSTM models whose hidden states only depend on the previous state.
    We can modify \ARC to record the previous $k$ string transformations instead of one.
    However, this modification will introduce an extra $O(\len{\bfx}^{k-1})$ time complexity in the algorithm.
    To avoid the additional time complexity, we ask: \emph{can we rewrite \cref{eq: def_trans} into equations such that the $n$th state only depend on the previous state?}

\paragraph{Our Approach}
The short answer to the above question is yes. In fact, \textbf{\emph{one-layer decoder-only transformers are two-layer RNNs.}}
\Cref{fig: illu_rewrite} and \Cref{sec: observation} illustrate this claim.

When designing the abstract transformer of \Cref{eq: rnn_trans}, we propose rewriting $f_i$ in two ways:
\begin{align*}
    f_i &= (v_i-f_{i-1})\sigma(q_nk_i - g_{i-1}) + f_{i-1}\\
    f_i &= v_i + (f_{i-1}-v_i)\sigma(g_{i-1} - q_nk_i)
\end{align*}
We then meet these two intervals as $\ibp{f}_i$.

\section{Evaluation}
We evaluate \ARCT using a single-layer, decoder-only transformer with two-head attention. 
Classification is made by feeding the final state of the transformer layer into two MLP layers.
We implemented a prototype of \ARCT on $\Tinsword{}$ and $\Tsubword{}$ to demonstrate the preliminary results\footnote{\url{https://anonymous.4open.science/r/cert_transformer/src/ibp.py}}.

\begin{table}[t]
    \centering
    \caption{Results of applying \ARCT to decoder-only Transformer on SST2 dataset. }
    \setlength{\tabcolsep}{3pt}

    \begin{tabular}{lrrr}
        \toprule
         & \multicolumn{3}{c}{$\{(\Tinsword{}, 2),(\Tsubword{},2)\}$} \\ 
         
         \cmidrule(lr){2-4} 
        Train & {Acc.} & {\Verifiedshort{}} & {\Exhaustiveshort{}}  \\
        \midrule
        Normal & 79.90 & 9.06 & 63.32\\
        Data Aug. & \textbf{80.51} & 8.95 & 67.33\\
        HotFlip & 79.79 & 14.22 & 71.22\\
        \ARCT & 79.19 & \textbf{64.80} & \textbf{72.71}\\
        \bottomrule
    \end{tabular}
    \label{tab:trans_res}
    \vspace{-1em}
\end{table}

We compare \ARCT with normal training, data augmentation, and HotFlip augmentation on the SST2 dataset against $\{(\Tinsword{}, 2),(\Tsubword{},2)\}$ using the following evaluation metrics.

\noindent \textbf{Normal accuracy} is the vanilla accuracy of the model on the test set.

\noindent \textbf{Certified accuracy} (\Verifiedshort{}) 
is the percentage of points in the test set
certified as $S$-robust (\Cref{eq:robustness}) using \ARCT.
    
\noindent \textbf{\Oracle{} accuracy}
is the worst-case accuracy of the model: a prediction on $(\bfx,y)$ is considered correct
if and only if all points $\bfz \in \spec(\bfx)$ lead to the correct prediction.
Formally, given a dataset $\dataset = \{(\bfx_i,y_i)\}_{i=1}^n$
and a perturbation space $\spec$, we define \oracle{} accuracy as follows:
\begin{align}
    \frac{1}{n} \sum_{i=1}^n \mathds{1}[\forall \bfz \in \spec(\bfx_i).\ \nnmodel_\params(\bfz)=y_i] \label{eq: exhaustiveacc}
\end{align}
Intuitively, for each sample $(\bfx_i,y_i)$, its classification is considered correct iff $\nnmodel_\params$ predicts $y_i$ for every single point in $\spec(\bfx_i)$.

\Cref{tab:trans_res} shows that \ARCT achieves the best certified accuracy and exhaustive accuracy when compared to the other three training approaches.

Compared to Table~4 in \citet{arc}, although the normal accuracy of our transformer model is lower than that of the LSTM model, the exhaustive accuracy and certified accuracy of the transformer model are higher. This suggests that the transformer model can exhibit stronger robustness or be more amenable to verification than the LSTM model.
Similar results have also been obtained by related work in randomized smoothing~\citep{text_crs}.

\clearpage
\section{Limitations and Future Work}
First, notice that $\sigma(q_nk_i - g_{i-1}) + \sigma(g_{i-1} - q_nk_i) = 1$ in \cref{eq: rnn_trans}, indicating that using the Zonotope abstract domain~\citep{transformer_lp_2} to replace the interval abstract domain is a promising future direction.

Second, unlike conventional transformer layers, we omit layer normalization in our evaluation, leaving the abstract transformer for layer normalization as a potential avenue for future research. Extending the evaluation to more perturbation spaces is also a future direction to further assess \ARCT's effectiveness.

Third, rewriting a one-layer decoder-only transformer into two RNNs slows down the attention mechanism computation. Two RNNs need to compute the final states sequentially, while the original attention computation in transformers leverages matrix multiplication, which is much faster than sequential computations of RNN states on GPUs.

Fourth, as large language models for code tasks become prevalent, it is crucial to design programmable code perturbation following the idea of programmable perturbation space to improve the robustness of code generation models~\citep{zhang2024codefort}.
The approaches for verifying decoder-only transformers proposed in this paper underlie the verification of these decoder-only-transformer-based large code models.

\newpage
\bibliography{iclr2024_conference}

\newpage
\appendix
\section{Proof of Theorem~\ref{theorem: rewrite}}

\begin{proof}
    We first prove the following equation of $g_i$ by induction.
    \begin{align}
        \mathrm{e}^{g_i} = \sum_{j=1}^i \mathrm{e}^{q_nk_j} \label{eq: rewrite_lemma1}
    \end{align}
    It is easy to see the base case ($i=1$) holds.
    In the inductive step, suppose $g_{i-1}$ holds for \Cref{eq: rewrite_lemma1}, we have 
\[
        \mathrm{e}^{g_i} = \mathrm{e}^{g_{i-1}} + \mathrm{e}^{q_nk_i} = \sum_{j=1}^{i-1} \mathrm{e}^{q_nk_j} + \mathrm{e}^{q_nk_i} = \sum_{j=1}^i \mathrm{e}^{q_nk_j}
\]

We then prove the following equation of $f_l$ by induction.
    \begin{align}
        f_l = \sum_{i=1}^l v_{i} \frac{\mathrm{e}^{q_nk_i}}{\sum_{j=1}^l \mathrm{e}^{q_nk_j}} \label{eq: rewrite_lemma2}
    \end{align}
    It is easy to see the base case ($l=1$) holds.
    In the inductive step, suppose $f_{l-1}$ holds for \Cref{eq: rewrite_lemma2}, we have 
    \begin{align*}
        f_l &= \sum_{i=1}^l v_{i} \frac{\mathrm{e}^{q_nk_i}}{\sum_{j=1}^l \mathrm{e}^{q_nk_j}}  \\
        &=v_l \frac{\mathrm{e}^{q_nk_l}}{\sum_{j=1}^l \mathrm{e}^{q_nk_j}} +  \sum_{i=1}^{l-1} v_{i} \frac{\mathrm{e}^{q_nk_i}}{\sum_{j=1}^l \mathrm{e}^{q_nk_j}}  \\
        &= v_{l} \frac{\mathrm{e}^{q_nk_l}}{\sum_{j=1}^{l-1} \mathrm{e}^{q_nk_j} + \mathrm{e}^{q_nk_l}} \\ & ~~~~+ \sum_{i=1}^{l-1} v_{i} \frac{\mathrm{e}^{q_nk_i}}{\sum_{j=1}^{l-1} \mathrm{e}^{q_nk_j}}\frac{\sum_{j=1}^{l-1} \mathrm{e}^{q_nk_j}}{\sum_{j=1}^{l-1} \mathrm{e}^{q_nk_j} + \mathrm{e}^{q_nk_l}} \\
        &= v_{l} \frac{\mathrm{e}^{q_nk_l}}{\mathrm{e}^{g_{l-1}} + \mathrm{e}^{q_nk_l}} \\ & ~~~~ + \sum_{i=1}^{l-1} v_{i} \frac{\mathrm{e}^{q_nk_i}}{\sum_{j=1}^{l-1} \mathrm{e}^{q_nk_j}}\frac{\mathrm{e}^{g_{l-1}}}{\mathrm{e}^{g_{l-1}} + \mathrm{e}^{q_nk_l}} \tag*{(By \Cref{eq: rewrite_lemma1})} \\
        &= v_{l} \frac{\mathrm{e}^{q_nk_l}}{\mathrm{e}^{g_{l-1}} + \mathrm{e}^{q_nk_l}} \\ & ~~~~ + f_{l-1}\frac{\mathrm{e}^{g_{l-1}}}{\mathrm{e}^{g_{l-1}} + \mathrm{e}^{q_nk_l}} \tag*{(By the inductive hypothesis)}\\
        &= v_l\sigma(q_nk_l - g_{l-1}) + f_{l-1}\sigma(g_{l-1} - q_nk_l)
    \end{align*}
    By the definition \Cref{eq: def_trans}, $\trans(\bfz_{1:n},m)$ is equivalent to $f_n$ in \Cref{eq: rewrite_lemma2}.
\end{proof}




\end{document}